\documentclass{article}

\usepackage[preprint,nonatbib]{neurips_2024}

\usepackage[numbers]{natbib}

\usepackage[utf8]{inputenc} 
\usepackage[T1]{fontenc}    
\usepackage{hyperref}       
\usepackage{url}            
\usepackage{booktabs}       
\usepackage{amsfonts}       
\usepackage{nicefrac}       
\usepackage{microtype}      
\usepackage{xcolor}         
\usepackage{amsmath}
\usepackage{cleveref}

\usepackage{mathtools}  
\usepackage{booktabs}  
\usepackage{multirow}  
\usepackage{listings}
\usepackage{xcolor}
\usepackage{hyperref}  

\newcommand{\bx}{\mathbf{x}}
\newcommand{\bM}{\mathbf{M}}
\newcommand{\R}{\mathbb{R}}
\newcommand{\N}{\mathcal{N}}
\newcommand{\ie}{\textit{i}.\textit{e}.}

\def\secref#1{Section~\ref{#1}}
\def\figref#1{Figure~\ref{#1}}
\def\tabref#1{Table~\ref{#1}}
\def\eqref#1{Eq.~\ref{#1}}

\DeclareMathOperator{\Dropout}{Dropout}

\DeclareMathOperator{\Bernoulli}{Bernoulli}
\DeclareMathOperator{\SSmy}{SS}
\DeclareMathOperator{\E}{E}
\DeclareMathOperator{\Var}{Var}
\DeclareMathOperator{\AP}{AP}
\DeclareMathOperator{\SAP}{SAP}

\definecolor{codegreen}{rgb}{0,0.6,0}
\definecolor{codegray}{rgb}{0.5,0.5,0.5}
\definecolor{codepurple}{rgb}{0.58,0,0.82}
\definecolor{backcolour}{rgb}{0.95,0.95,0.92}

\lstdefinestyle{mystyle}{
    backgroundcolor=\color{backcolour},
    commentstyle=\color{codegreen},
    keywordstyle=\color{magenta},
    numberstyle=\tiny\color{codegray},
    stringstyle=\color{codepurple},
    basicstyle=\ttfamily\footnotesize,
    breakatwhitespace=false,
    breaklines=true,
    captionpos=b,
    keepspaces=true,
    numbers=left,
    numbersep=5pt,
    showspaces=false,
    showstringspaces=false,
    showtabs=false,
    tabsize=2
}

\title{Stochastic Subsampling With Average Pooling}

\author{%
  Bum Jun Kim \\
  POSTECH \\
  \texttt{kmbmjn@postech.edu} \\
  \And
  Sang Woo Kim \\
  POSTECH \\
  \texttt{swkim@postech.edu} \\
}

\begin{document}

\maketitle

\begin{abstract}
	Regularization of deep neural networks has been an important issue to achieve higher generalization performance without overfitting problems. Although the popular method of Dropout provides a regularization effect, it causes inconsistent properties in the output, which may degrade the performance of deep neural networks. In this study, we propose a new module called stochastic average pooling, which incorporates Dropout-like stochasticity in pooling. We describe the properties of stochastic subsampling and average pooling and leverage them to design a module without any inconsistency problem. The stochastic average pooling achieves a regularization effect without any potential performance degradation due to the inconsistency issue and can easily be plugged into existing architectures of deep neural networks. Experiments demonstrate that replacing existing average pooling with stochastic average pooling yields consistent improvements across a variety of tasks, datasets, and models.
\end{abstract}

\section{Introduction}

Deep neural networks have demonstrated remarkable capabilities in a variety of fields, such as computer vision and natural language processing benchmarks \cite{DBLP:journals/nn/ChengSQCC24,DBLP:journals/nn/LanHWLLMNL24,DBLP:journals/nn/AhmadianRFO24}. With a large number of parameters, they are able to represent abstract and high-level patterns in data, which has led to significant improvements in modeling abilities. Despite their successes, the large number of parameters often becomes over-parameterized, which causes overfitting problems to the training dataset and thereby degrades the generalization performance \cite{DBLP:journals/nn/HeidariMG23,DBLP:journals/nn/LiYH23}. To avoid the overfitting problem, current practices in deep learning necessitate sufficient regularization methods, such as weight decay \cite{DBLP:conf/iclr/ZhangWXG19}, normalization layers \cite{DBLP:conf/icml/IoffeS15}, and Dropout \cite{DBLP:journals/jmlr/SrivastavaHKSS14}.

Dropout is a famous regularization method adopted for deep neural networks. Dropout turns off arbitrary neurons within a neural network during the training phase, which enables training of a subnetwork that is randomly sampled. During the test phase, the whole network is used for inference, which becomes an ensemble of all possible subnetworks. This ensemble behavior yields a regularization effect on the deep neural network, which alleviates overfitting problems.

However, the recent practice of using batch normalization \cite{DBLP:conf/icml/IoffeS15} raises a side effect from Dropout. Batch normalization expects consistent mean and variance in its input, whereas Dropout causes inconsistent variance during training and test phases \cite{DBLP:conf/cvpr/0028C0019}. In other words, the use of Dropout breaks the fundamental assumption of batch normalization, which degrades performance when used together. Reference \cite{DBLP:conf/uai/KimCJLK23} proved that this inconsistency problem cannot be avoided for any variant of the Dropout scheme but could be partially mitigated by adopting indirect means such as choosing a proper position for Dropout.

In this study, we explore a form of alternative module that achieves a Dropout-like regularization effect without introducing the side effect of the inconsistency issue. Here, we extend and generalize the recent operation of PatchDropout \cite{DBLP:conf/eccv/LinMBHPRDZ14,DBLP:conf/cvpr/HeCXLDG22,DBLP:conf/cvpr/Li0HFH23} as stochastic subsampling and merge it into average pooling with an adequate scaling factor. To this end, we propose \emph{stochastic average pooling}, which enables the average pooling to obtain stochasticity. The stochastic average pooling provides a Dropout-like regularization effect and ensures consistency in output properties. The existing average pooling can be seamlessly replaced with stochastic average pooling by inserting few lines of code. Experiments showed that replacing average pooling with stochastic average pooling improved performance across numerous datasets, tasks, and models.

\section{Method}

\subsection{Preliminaries: Dropout and PatchDropout}

\begin{table}[t!]
	\centering
	\resizebox{1.0\textwidth}{!}{
		\begin{tabular}{ll}
			\toprule
			\textbf{Notation}        & \textbf{Meaning}                                                               \\
			\midrule
			$\bx$                    & A vector $\bx=[x_1, x_2, \cdots, x_n]$ of length $n$.                          \\
			$p$                      & Keep probability $p \in (0, 1)$.                                               \\
			$\bM$                    & An $n \times n$ diagonal matrix with Bernoulli variable.                       \\
			$\E[\bx]$                & Mean of vector $\bx$.                                                          \\
			$\E[\bx^2]$              & Mean of vector $\bx^2$, \ie, second moment of $\bx$.                           \\
			$\Var[\bx]$              & Variance of vector $\bx$.                                                      \\
			$\SSmy$                  & Stochastic subsampling operation.                                              \\
			$\AP^r$                  & An $r$-size 1D average pooling with stride $r$.                                \\
			$\AP^{r \times r}$       & An $(r \times r)$-size 2D average pooling with stride $r \times r$.            \\
			$\SAP^r$                 & An $r$-size 1D stochastic average pooling with stride $r$.                     \\
			$\SAP^{r \times r}$      & An $(r \times r)$-size 2D stochastic average pooling with stride $r \times r$. \\
			$k_i$                    & Probability map used in stochastic pooling.                                    \\
			$\N(0, 1)$               & Standard normal distribution with zero mean and unit variance.                 \\
			$N, C, H, W$             & Mini-batch size, the number of channels, height, and width.                    \\
			$l \times l, s \times s$ & Sizes of input and factor for design of spatial pattern.                       \\
			\bottomrule
		\end{tabular}
	}
	\caption{List of notations used in this study}
	\label{tab:notation}
\end{table}

The full list of mathematical notations used in this study is summarized in \tabref{tab:notation}. Let $\bx=[x_1, x_2, \cdots, x_n]$ be a vector of length $n$. With a keep probability $p \in (0, 1)$, Dropout randomly drops certain elements from the vector during the training phase, whereas it passes intact elements without any drop during the test phase \cite{DBLP:journals/jmlr/SrivastavaHKSS14}. This standard definition of Dropout can be described as
\begin{align}
	\Dropout_\textrm{train}(\bx) & \coloneqq \frac{1}{p} \bM \bx, \\
	\Dropout_\textrm{test}(\bx)  & \coloneqq \bx,
\end{align}
where $\bM$ is an $n \times n$ diagonal matrix with $m_{ij}=0$ for $i \neq j$ and $m_{ij} \sim \Bernoulli(p)$ for $i=j$. Note that Dropout randomly converts certain elements in the vector into zeros, whereas others are left with $1/p$ scaling, yielding a vector such as $[0, x_2/p, \cdots, 0]$. A Dropout-applied neural network behaves as an ensemble of its subnetworks, which acquires a regularization effect to avoid an overfitting problem. Since the introduction of Dropout, a keep probability such as $p=0.5$ or $p=0.8$ has been preferred in the research community \cite{DBLP:journals/jmlr/SrivastavaHKSS14,DBLP:journals/corr/SimonyanZ14a,DBLP:conf/icml/TanL19,DBLP:conf/icml/BrockDSS21}.

Inspired by Dropout, recent studies on vision transformers have employed its variant called PatchDropout \cite{DBLP:conf/eccv/LinMBHPRDZ14,DBLP:conf/cvpr/HeCXLDG22,DBLP:conf/cvpr/Li0HFH23}. In vision transformers \cite{DBLP:conf/iclr/DosovitskiyB0WZ21}, a 2D image is partitioned into patches to represent it as a sequence of features on patches, which is referred to as patch embeddings. PatchDropout randomly subsamples certain patches of patch embeddings in the early stage of the vision transformer during the training phase. Removing other unchosen patches saves computational resources, thereby speeding up the training of the vision transformer. During the test phase, PatchDropout passes the patch embeddings without any subsampling. Indeed, several pieces of literature \cite{DBLP:conf/eccv/LinMBHPRDZ14,DBLP:conf/cvpr/HeCXLDG22,DBLP:conf/cvpr/Li0HFH23} reported that by adopting a low keep probability such as $p=0.5$ or $p=0.25$, applying PatchDropout into a vision transformer leads to two to three times faster training speeds while maintaining accuracy.

However, PatchDropout has only been used in this limited scenario of subsampling the patch embeddings of a vision transformer. To extend and generalize the usage of PatchDropout for general purposes, here we define \emph{stochastic subsampling} (SS) as
\begin{align}
	\SSmy_\textrm{train}(\bx) & \coloneqq [x_i]_{i \in S_p}, \label{eq:sstrain} \\
	\SSmy_\textrm{test}(\bx)  & \coloneqq \bx, \label{eq:sstest}
\end{align}
where a set $S_p$ consists of element indices of a randomly selected subset from $[1, 2, \cdots, n]$ in length $np$ without duplication. The random indices change every time.

Although PatchDropout---or stochastic subsampling---may seem similar to Dropout, in fact, they differ in several points. For Dropout, the dropped elements are regarded as zeros, whereas PatchDropout simply removes them by subsampling. In a strict sense, PatchDropout is not a Dropout. To clarify the difference between them and generalize PatchDropout beyond its current usage for patch embedding, in this study, we use the term stochastic subsampling.

In the Dropout scheme, to alleviate decreased mean due to zeroed elements, a scaling factor of $1/p$ is employed during the training phase. This scaling ensures mean consistency during training and test phases: $\E[\frac{1}{p}\bM \bx] = \E[\bx]$. Nevertheless, the problem we claim is that zeroed elements by Dropout affect the mean of the squares for an output, \ie, the second moment:
\begin{align}
	\E[\frac{1}{p^2} m_{i,i}^2 x_i^2] = \frac{1}{p}\E[x_i^2] > \E[x_i^2],
\end{align}
thereby causing increased variance during the training phase compared with variance during the test phase \cite{DBLP:conf/cvpr/0028C0019}. Reference \cite{DBLP:conf/uai/KimCJLK23} proved that Dropout cannot simultaneously satisfy both mean and variance consistency, even for other possible variants such as different choices for the distribution of $\bM$. The inconsistency in mean or variance breaks the underlying assumption of subsequent batch normalization: Though batch normalization expects to receive features of the same mean and variance during training and test phases, Dropout causes inconsistency in mean or variance. Owing to this problem, the use of Dropout with batch normalization may degrade performance and has been avoided in the research community \cite{DBLP:conf/cvpr/0028C0019,DBLP:conf/icml/IoffeS15,DBLP:conf/cvpr/HeZRS16}. Even though Reference \cite{DBLP:conf/uai/KimCJLK23} found that applying Dropout at a specific position partially resolves the inconsistency, they concluded that the inconsistency issue always exists in the Dropout scheme.

\begin{figure}[t!]
	\centering
	\includegraphics[width=1.0\textwidth]{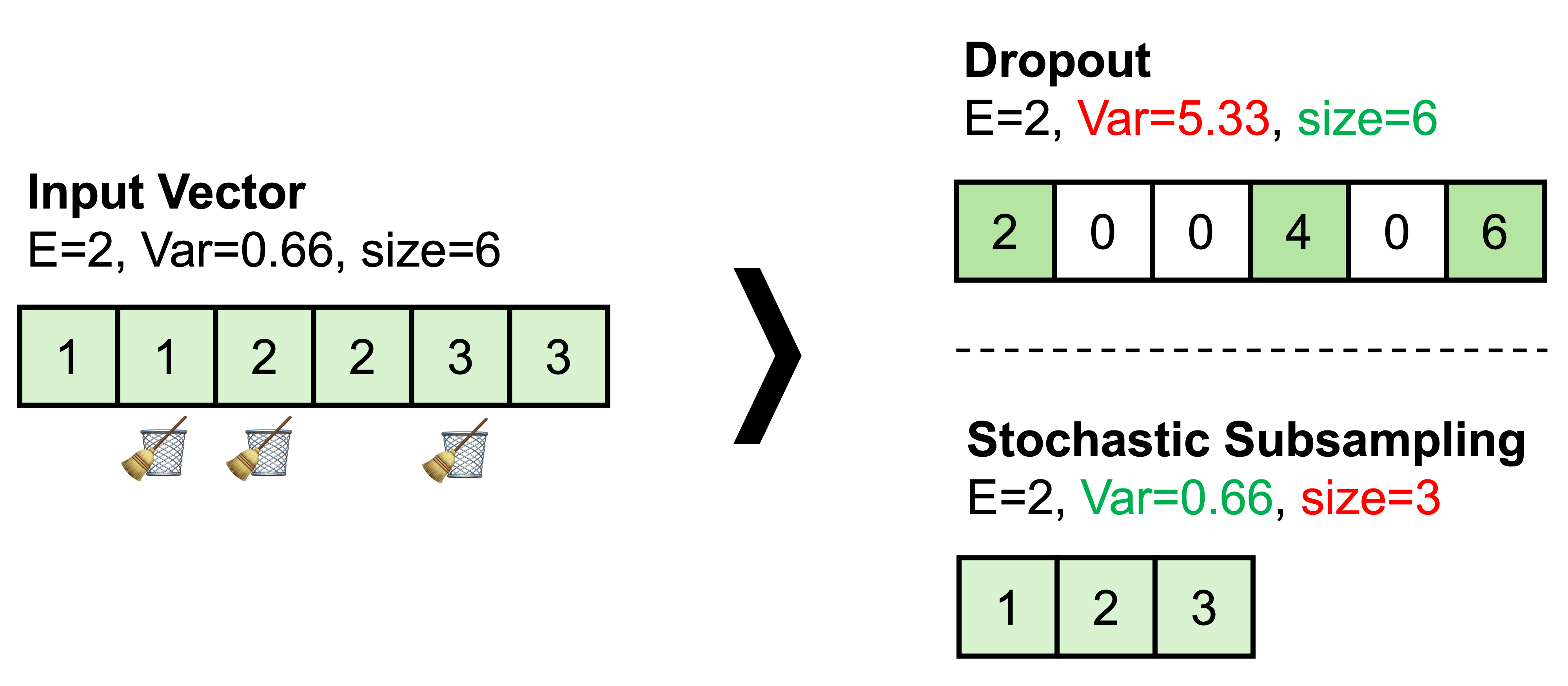}
	\caption{Illustration of Dropout and stochastic subsampling for a keep probability of $p=0.5$. Dropout erases half of the elements into zeros and scales the vector by $1/p$, causing increased variance. Stochastic subsampling yields a subvector of the input, which conserves variance. Nevertheless, stochastic subsampling reduces the size of the vector.}
	\label{fig:dropout_ss}
\end{figure}

In contrast, we find that stochastic subsampling does not introduce zeroed elements (\figref{fig:dropout_ss}), which guarantees consistency in both the mean and variance during training and test phases:
\begin{align}
	\E[\SSmy_\textrm{train}(\bx)]   & = \E[\SSmy_\textrm{test}(\bx)],   \\
	\Var[\SSmy_\textrm{train}(\bx)] & = \Var[\SSmy_\textrm{test}(\bx)].
\end{align}
These consistencies in mean and variance hold when sampling a subset from a vector with a sufficiently large size, such as a feature map in deep neural networks. Considering the consistencies in mean and variance, stochastic subsampling is a much safer choice to be deployed for deep neural networks with batch normalization, compared with Dropout.

However, stochastic subsampling reduces the size of the vector from $n$ into $np$ during the training phase, whereas it maintains the vector size $n$ during the test phase. Nevertheless, the architecture of a deep neural network may require a consistent size for an intermediate feature map. For example, a fully connected (FC) layer requires a fixed size of input vector, whereas stochastic subsampling changes the vector size during training and test phases. Therefore, stochastic subsampling cannot be deployed with an FC layer. Owing to its size reduction, stochastic subsampling has only been used in the limited scenario of the early stage of vision transformers as PatchDropout, and its general usage has been rarely studied up to now.

\subsection{Proposed Method: Stochastic Average Pooling}
\label{sec:sap}

The objective of this study is to design a module using stochastic subsampling to achieve a Dropout-like regularization effect as well as consistent vector properties such as variance and size. To cope with the reduced size of a vector, we exploit the current practice of using average pooling, which behaves as downsampling for image features. Here, we investigate a form of new module using average pooling that incorporates stochastic subsampling.

We first start with the standard definition of average pooling. Given a vector $\bx = [x_1, x_2, \cdots, x_n]$, the $j$th element from $r$-size 1D average pooling with stride $r$ is
\begin{align}
	\AP^r_{j}(\bx) \coloneqq \frac{1}{r} \sum_{i} x_{i}, \label{eq:ap}
\end{align}
where $i \in \{r(j-1)+1, r(j-1)+2, \cdots, r(j-1) + r\}$ to cover the specified pooling size. Here, we find that average pooling conserves the mean but not the variance, owing to the decreased second moment
\begin{align}
	\E[\AP^r_{j}(\bx)^2] \approx \frac{1}{r^2} \sum_{i} x_i^2 = \frac{1}{r} \E[x_i^2]. \label{eq:inverse}
\end{align}
In other words, applying $r$-size 1D average pooling with stride $r$ decreases the second moment by $1/r$, approximately. This behavior will be empirically verified in \secref{sec:empi}. Global average pooling (GAP) \cite{DBLP:journals/corr/LinCY13} corresponds to its special form of $r=n$, which decreases the second moment by a factor of $1/n$. Similarly, $(r \times r)$-size 2D average pooling with stride $r \times r$, denoted by $\AP^{r \times r}$, decreases the second moment by $1/r^2$. In summary, pooling size $r$ determines the decreased second moment after average pooling, which should be considered when incorporating stochastic subsampling.

\begin{figure}[t!]
	\centering
	\includegraphics[width=1.0\textwidth]{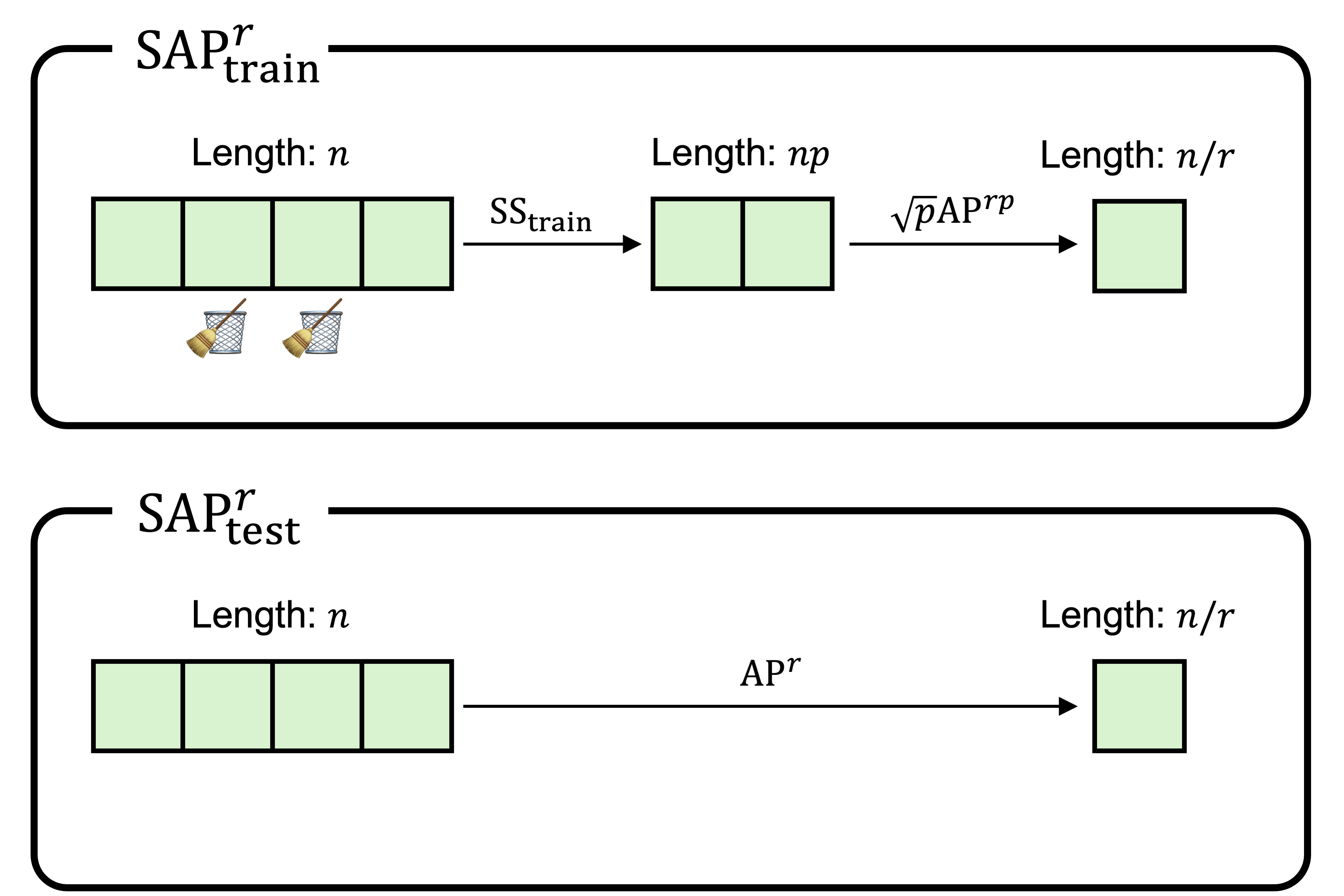}
	\caption{Illustration of stochastic average pooling during training and test phases for a 1D vector}
	\label{fig:sap1d}
\end{figure}

\begin{figure}[t!]
	\centering
	\includegraphics[width=1.0\textwidth]{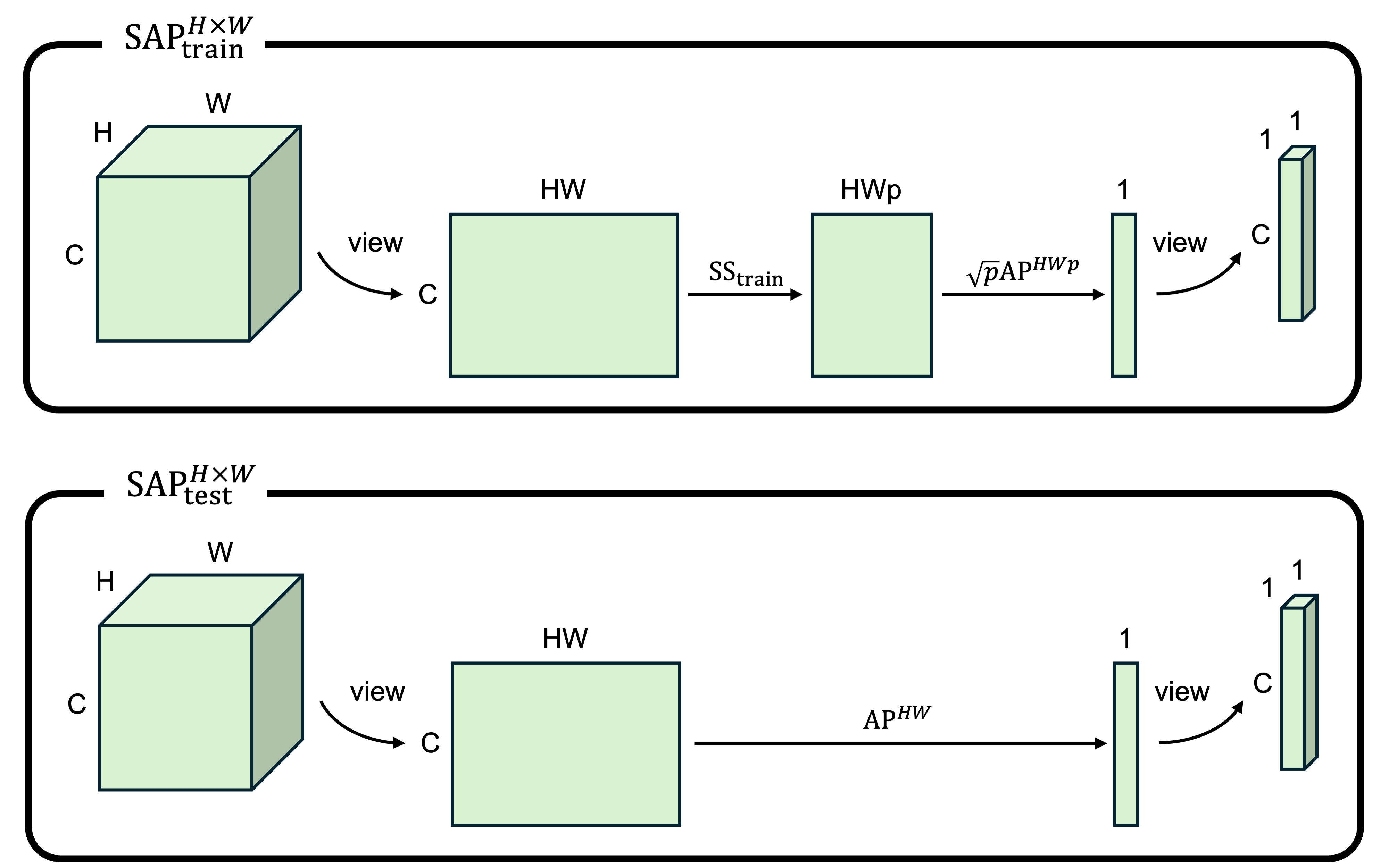}
	\caption{Illustration of stochastic average pooling for 2D image feature. We flatten the spatial dimensions and apply stochastic average pooling to each channel.}
	\label{fig:sap2d}
\end{figure}

Now, we combine average pooling and stochastic subsampling. Firstly, we define a vanilla $r$-size 1D average pooling with stride $r$, which is going to be used during the test phase. This average pooling reduces the size of a vector from $n$ to $n/r$. For use during the training phase, we consider stochastic subsampling that is followed by another average pooling operation. Because the stochastic subsampling reduces the size of a vector from $n$ to $np$, here we employ $(rp)$-size 1D average pooling with stride $rp$ that reduces the size of a vector from $np$ to $n/r$. Note that, though stochastic subsampling conserves the second moment, the subsequent average pooling affects the second moment. Specifically, $(rp)$-size 1D average pooling with stride $rp$ decreases the second moment by a factor of $1/rp$ during the training phase, whereas $r$-size 1D average pooling with stride $r$ causes decreased second moment by a factor of $1/r$ during the test phase. To match second moments that differ by a factor of $p$, we claim to apply $\sqrt{p}$ scaling during the training phase. In summary, we propose a new module, \emph{stochastic average pooling} (SAP), as
\begin{align}
	\SAP^r_{\textrm{train}}(\bx) & \coloneqq \sqrt{p} \AP^{rp}(\SSmy_{\textrm{train}}(\bx)), \label{eq:saptrain} \\
	\SAP^r_{\textrm{test}}(\bx)  & \coloneqq \AP^r(\bx),  \label{eq:saptest}
\end{align}
where $p \in (0, 1)$ corresponds to the keep probability of inner stochastic subsampling. Given a vector $\bx$ with size $n$, this definition of stochastic average pooling consistently outputs a vector with a size of $n/r$ during both training and test phases. Furthermore, the $\sqrt{p}$ scaling calibrates the decreased second moment due to different average poolings, thereby matching them during both training and test phases. Notably, $\sqrt{p}$ scaling leads to amplified mean; Nevertheless, we empirically observed that applying additional mean calibration was unnecessary and rather slightly degraded performance. Hence, we opt for the above design of stochastic average pooling. The above definition of stochastic average pooling corresponds to the 1D case (\figref{fig:sap1d}); For computer vision tasks, we adopt its 2D version $\SAP^{r \times r}$ (\figref{fig:sap2d}).

\begin{figure}[t!]
	\centering
	\includegraphics[width=1.0\textwidth]{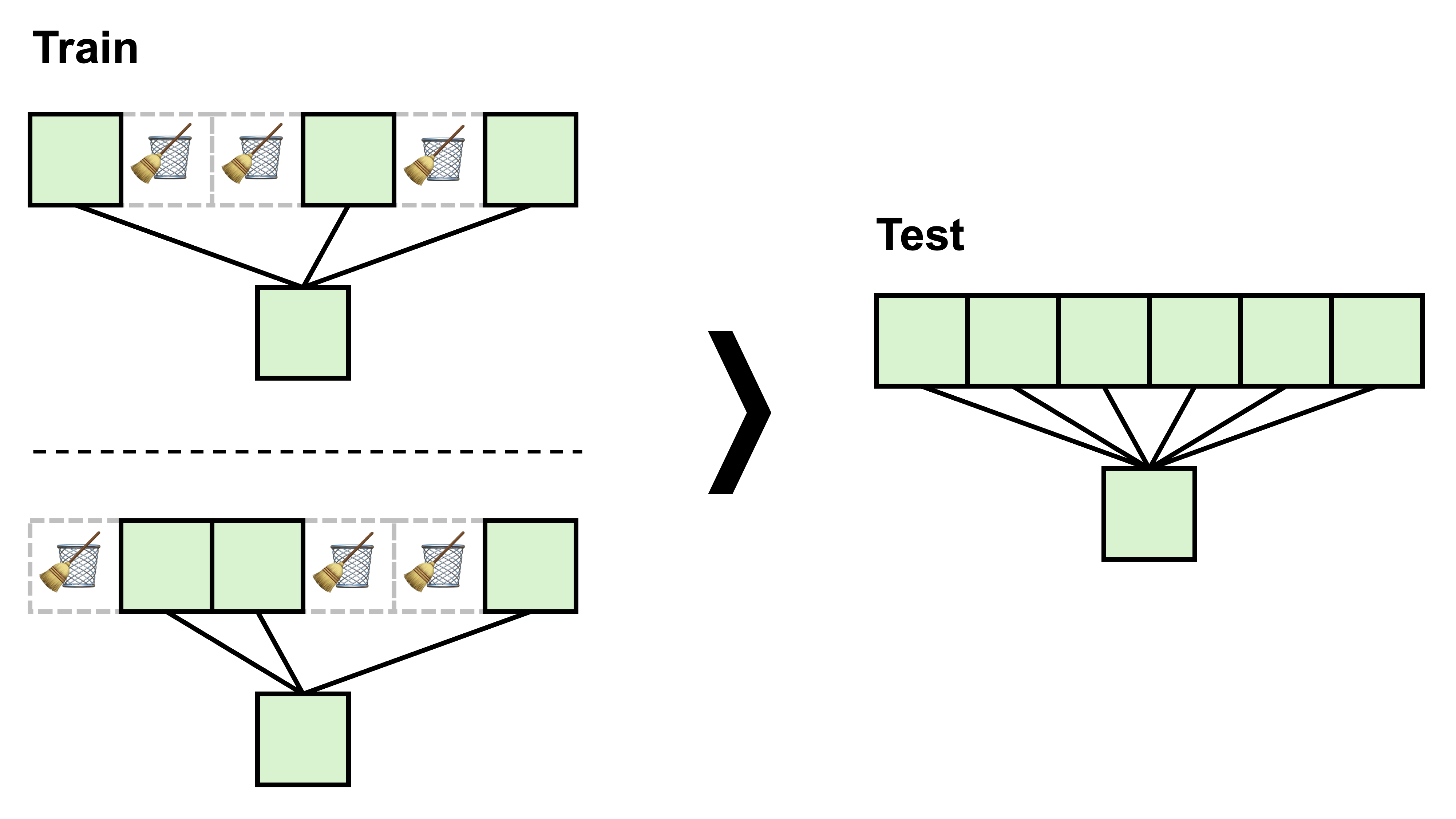}
	\caption{During the training phase, stochastic average pooling behaves as average pooling on a subnetwork that is randomly sampled. During the test phase, it operates as vanilla average pooling that becomes an ensemble of all possible subnetworks.}
	\label{fig:ensemble}
\end{figure}

During the training phase, stochastic average pooling combines arbitrarily selected elements on input by stochastic subsampling, which behaves as a subnetwork that is randomly sampled. During the test phase, stochastic average pooling operates as vanilla average pooling over the elements within the specified pooling size. The vanilla average pooling becomes an ensemble of all possible subnetworks during the training phase, which obtains a regularization effect similar to Dropout (\figref{fig:ensemble}).

In summary, stochastic average pooling embraces stochastic subsampling during the training phase, whereas it behaves as vanilla average pooling during the test phase. Leveraging this behavior, the existing average pooling used in the architecture of deep neural networks can be replaced with stochastic average pooling to introduce an additional Dropout-like regularization effect during training. Indeed, average pooling has been widely used in a variety of tasks using deep neural networks. Firstly, most classification networks such as residual networks \cite{DBLP:conf/cvpr/HeZRS16} have used a classifier head that employs global average pooling and an FC layer. In addition, global average pooling has been deployed in channel-wise attention, such as the squeeze-and-excitation (SE) block \cite{DBLP:conf/cvpr/HuSS18} and its variants \cite{DBLP:conf/cvpr/LiW0019,DBLP:conf/cvpr/WangWZLZH20}, to obtain aggregated information for each channel. Moreover, average pooling has been widely used in various computer vision models such as PSPNet \cite{DBLP:conf/cvpr/ZhaoSQWJ17} and UPerNet \cite{DBLP:conf/eccv/XiaoLZJS18} to fuse multi-level feature maps in global and local contexts. Finally, recent architectural design patterns \cite{DBLP:conf/cvpr/HeZ0ZXL19,DBLP:conf/icml/BrockDSS21,DBLP:conf/eccv/TuTZYMBL22} prefer average pooling to max pooling to prevent information bottleneck. All of these usages belong to prime examples where stochastic average pooling becomes a better alternative than vanilla average pooling.

\begin{lstlisting}[language=Python, caption=Python example for the implementation of 2D stochastic average pooling., label=list:torch, float=t]
import torch
import torch.nn as nn
import math


class StochasticAvgPool(nn.Module):
    def __init__(self):
        super(StochasticAvgPool, self).__init__()

    def forward(self, x, keep_prob=0.5):
        if self.training:
            N, C, H, W = x.size()
            x = x.view(N, C, -1)

            noise = torch.rand(N, H * W, device=x.device)
            ids_shuffle = torch.argsort(noise, dim=1)
            len_keep = int(H * W * keep_prob)
            ids_keep = ids_shuffle[:, :len_keep]
            ids_index = ids_keep.unsqueeze(1).repeat(1, C, 1)
            x = torch.gather(x, dim=2, index=ids_index)

            x = x.mean(dim=-1)
            x = math.sqrt(keep_prob) * x
            return x
        else:
            N, C, _, _ = x.size()
            return x.view(N, C, -1).mean(dim=-1)
\end{lstlisting}

To help readers easily deploy stochastic average pooling, we provide a Python implementation example of stochastic average pooling (Listing~\ref{list:torch}) using the PyTorch library \cite{DBLP:conf/nips/PaszkeGMLBCKLGA19}. This source code implements 2D stochastic average pooling for a feature map. Lines 15 to 20 correspond to the inner stochastic subsampling, which outputs a subset of randomly selected elements. This stochastic subsampling operation generates random noise to obtain randomized indices without duplication, similar to the implementation of PatchDropout \cite{DBLP:conf/eccv/LinMBHPRDZ14,DBLP:conf/cvpr/HeCXLDG22,DBLP:conf/cvpr/Li0HFH23}. The random indices are used for sampling a subset according to \eqref{eq:sstrain}. Subsequently, average pooling and $\sqrt{p}$ scaling are applied, as described in \eqref{eq:saptrain}. During the test phase, it simply operates as vanilla average pooling, following \eqref{eq:saptest}. Note that in Lines 22 and 27, we adopt an average pooling implementation with merged spatial dimension, which is suggested in the fast global average pooling of TResNet \cite{DBLP:conf/wacv/RidnikLNBSF21} to optimize operations on GPU resources. After applying this module definition, average pooling in the existing source code of deep neural networks can be easily replaced with stochastic average pooling by modifying few lines of code.

A close research to stochastic average pooling of this study is the stochastic pooling by Reference \cite{DBLP:journals/corr/abs-1301-3557}. They proposed to compute a probability map for a given vector and subsample it based on the probability map. Given a vector $\bx$, the probability map is computed by its normalized values $k_i \coloneqq x_i / \sum_i x_i$. The probability map determines sampling probability on each element and enables to subsample stronger elements in the vector. During the test phase, their stochastic pooling becomes the weighted average of the elements using the probability map, $\sum_i k_i x_i$. Although this approach similarly achieves a Dropout-like regularization effect in pooling, its behavior during the test phase differs from vanilla average pooling. In contrast, our stochastic average pooling behaves as a vanilla average pooling during the test phase and can seamlessly replace existing average pooling while introducing a Dropout-like regularization effect during the training phase. Furthermore, contrary to our stochastic average pooling, their stochastic pooling exhibits inconsistency in the second moment during training and test phases: $\sum_i k_i x_i^2 \neq (\sum_i k_i x_i)^2$, which may degrade performance when using batch normalization. Considering these differences, our stochastic average pooling becomes a more practical method to introduce stochasticity in pooling.

In summary, the desired objective is satisfied by our proposed scheme with a pipeline of stochastic subsampling, average pooling, and $\sqrt{p}$ scaling. However, several technical details should be considered. Specifically, in stochastic subsampling, different ways of selecting element indices could be allowed. For example, Dropout independently masks elements on a given vector, whereas PatchDropout subsamples certain patches using a mask that is shared across the channel dimension. In computer vision tasks, the latter way masks semantic information on target patches within an image, which drives a stronger regularization effect compared with the Dropout-like independent mask. Considering this property, we opt for the latter way of selecting element indices. Furthermore, recent studies such as DropBlock \cite{DBLP:conf/aaai/PhamL21} and AutoDropout \cite{DBLP:conf/nips/GhiasiLL18} reported improved performance via restriction in spatial pattern of Dropout, which is worthy of investigation. These design choices of subsampling indices will be studied thoroughly in \secref{sec:disc}.

\subsection{Empirical Observation}
\label{sec:empi}

Here, we empirically validate that applying $\sqrt{p}$ scaling in stochastic average pooling guarantees consistency in the second moment. We simulate a scenario of applying 2D global average pooling to a feature map $\bx$ that is randomly sampled from $\N(0, 1)$. We used $\bx \in \R^{N \times C \times H \times W}$, where we set the mini-batch size to $N=64$, the number of channels to $C=256$, and the spatial size to $(H \times W) \in \{2^2, 4^2, 8^2, 16^2, 32^2, 64^2, 128^2, 256^2\}$. To obtain the behavior of global average pooling, the pooling size is set to be equal to the spatial size $r \times r = H \times W$. The keep probability is set to $p=0.5$ here, and other different choices will be studied later. We measured the second moment after applying (1) stochastic average pooling during the test phase, \ie, global average pooling from \eqref{eq:saptest}, (2) stochastic average pooling during the training phase from \eqref{eq:saptrain}, and (3) stochastic average pooling during the training phase from \eqref{eq:saptrain} omitting $\sqrt{p}$ scaling.

\begin{figure}[t!]
	\centering
	\includegraphics[width=0.9\textwidth]{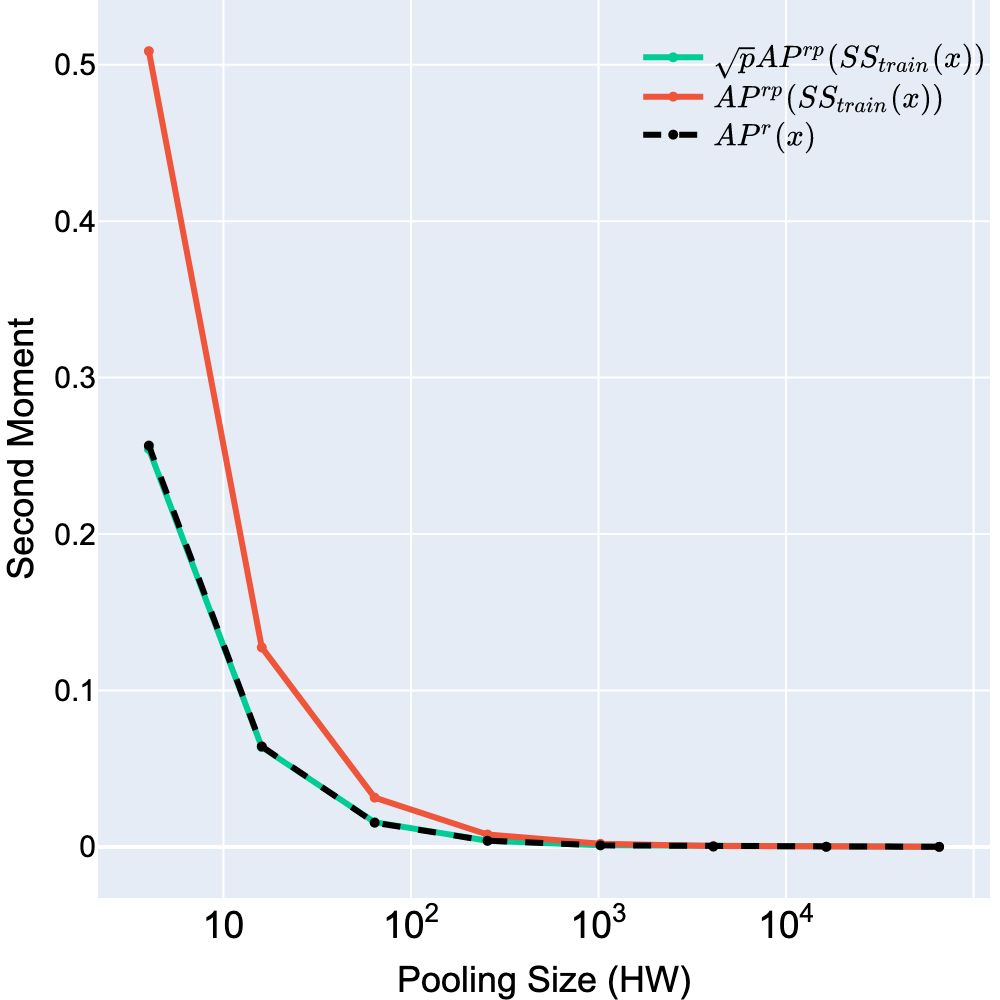}
	\caption{Simulation results for the second moment after stochastic average pooling. To obtain a consistent second moment during training and test phases, $\sqrt{p}$ scaling should be applied.}
	\label{fig:secondmomentum}
\end{figure}

The results are summarized in \figref{fig:secondmomentum}. We observed that the second moment after stochastic average pooling from \eqref{eq:saptrain} matched suitably with that after global average pooling from \eqref{eq:saptest}. If $\sqrt{p}$ scaling is omitted, then the second moment differed from that of global average pooling, which verifies the necessity of applying the $\sqrt{p}$ scaling. We emphasize that the different second moments arise from the distinct pooling sizes of the average poolings, whereas the stochastic subsampling conserves the second moment. We also observed that the second moment after global average pooling was in inverse proportion to the spatial size $HW$, which validates the decreased second moment after average pooling in \eqref{eq:inverse}.

\begin{figure}[t!]
	\centering
	\includegraphics[width=0.9\textwidth]{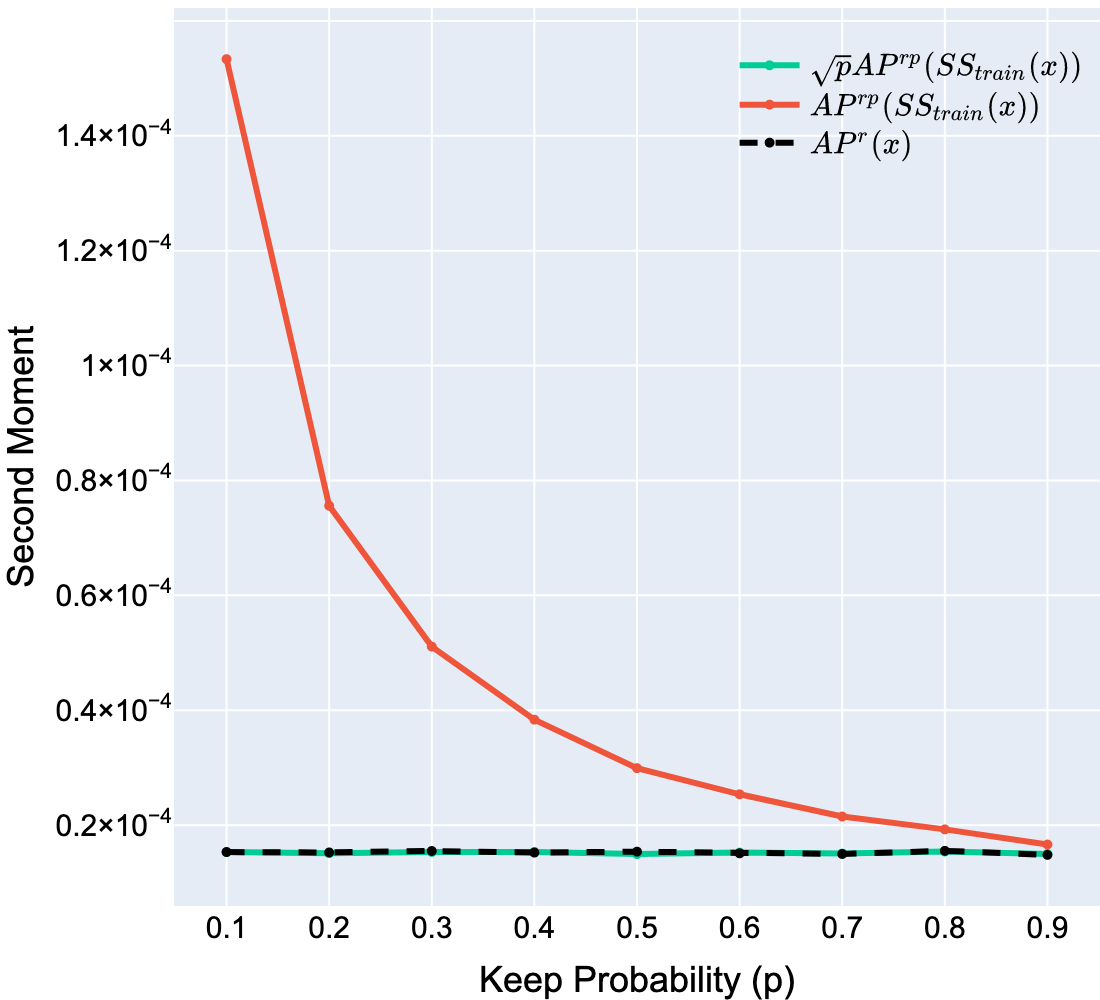}
	\caption{Simulation results for different setups regarding keep probability. Again, applying $\sqrt{p}$ scaling equalizes the second moment during training and test phases.}
	\label{fig:secondmomentum_p}
\end{figure}

Additionally, using the above setup, we investigate different choices of the keep probability $p \in \{0.1, 0.2, \cdots, 0.9\}$ for $H \times W = 256^2$. \figref{fig:secondmomentum_p} summarizes the simulation results. Again, we observed that applying $\sqrt{p}$ scaling ensured a consistent second moment for both stochastic average pooling and global average pooling for any choices of the keep probability, whereas omitting $\sqrt{p}$ caused inconsistency. All these simulation results validate our findings from the previous subsection.

\section{Experiments}

\begin{table}[t!]
	\centering
	\resizebox{1.0\textwidth}{!}{
		\begin{tabular}{lllll}
			\toprule
			\textbf{Dataset}  & \textbf{Reference}                    & \textbf{Task}         & \textbf{Website}                                              \\
			\midrule
			CIFAR-\{10, 100\} & \cite{krizhevsky2009learning}         & Image classification  & \url{https://www.cs.toronto.edu/~kriz/cifar.html}             \\
			Oxford-IIIT Pet   & \cite{DBLP:conf/cvpr/ParkhiVZJ12}     & Image classification  & \url{https://www.robots.ox.ac.uk/~vgg/data/pets/}             \\
			Caltech-101       & \cite{DBLP:journals/cviu/Fei-FeiFP07} & Image classification  & \url{http://www.vision.caltech.edu/datasets/}                 \\
			Stanford Cars     & \cite{DBLP:conf/iccvw/Krause0DF13}    & Image classification  & \url{https://ai.stanford.edu/~jkrause/cars/car_dataset.html}  \\
			ISPRS Potsdam     & \cite{rottensteiner2012isprs}         & Semantic segmentation & \url{https://www.isprs.org/education/benchmarks/UrbanSemLab/} \\
			ISPRS Vaihingen   & \cite{rottensteiner2012isprs}         & Semantic segmentation & \url{https://www.isprs.org/education/benchmarks/UrbanSemLab/} \\
			COCO 2017         & \cite{DBLP:conf/eccv/LinMBHPRDZ14}    & Object detection      & \url{https://cocodataset.org/\#home}                          \\
			\bottomrule
		\end{tabular}
	}
	\caption{List of datasets used for experiments. All of these datasets are publicly available.}
	\label{tab:dataset}
\end{table}

In this section, we experiment with numerous deep neural networks before and after replacing average pooling with stochastic average pooling. We target a variety of datasets, tasks, and models (\tabref{tab:dataset}), where average pooling has been deployed.

\subsection{Replace GAP in Classifier Head}
\label{sec:cifar}

Firstly, we examine the performance differences when replacing the existing global average pooling in a classifier head with stochastic average pooling. We trained ResNet-\{50, 110\} \cite{DBLP:conf/cvpr/HeZRS16} on a multi-class classification task using the CIFAR-\{10, 100\} dataset \cite{krizhevsky2009learning}. The CIFAR-\{10, 100\} dataset consists of 60K images of \{10, 100\} classes. For data augmentation, we used $32 \times 32$ random cropping with 4-pixel padding, a random horizontal flip with a probability of 0.5, and mean-std normalization using dataset statistics. For training, the number of epochs of 164, stochastic gradient descent with a momentum of 0.9, learning rate of 0.1, learning rate decay of 0.1 at \{81, 122\} epochs, weight decay of $10^{-4}$, and mini-batch size of 128 were used. The training was conducted on a single GPU machine. We examined different choices of keep probability $p \in \{0.1, 0.2, \cdots, 0.9\}$, and an average of ten runs with different random seeds was reported for each result (Tables~\ref{tab:c10}, \ref{tab:c100}).

\begin{table}[t!]
	\centering
	\begin{tabular}{l|rr|rr}
		\toprule
		\textbf{Dataset} & \multicolumn{4}{c}{\textbf{CIFAR-10}}                                                                                              \\
		\textbf{Model}   & \multicolumn{2}{c|}{\textbf{ResNet-50}} & \multicolumn{2}{c}{\textbf{ResNet-110}}                                                  \\
		\textbf{Setup}   & \textbf{Accuracy}                       & \textbf{Difference}                     & \textbf{Accuracy} & \textbf{Difference}        \\
		\midrule
		GAP (Baseline)   & 93.160                                  & -                                       & 93.639            & -                          \\
		SAP with $p=0.1$ & 93.010                                  & (\textcolor{red}{-0.150})               & 93.749            & (\textcolor{blue}{+0.110}) \\
		SAP with $p=0.2$ & 93.131                                  & (\textcolor{red}{-0.029})               & 93.821            & (\textcolor{blue}{+0.182}) \\
		SAP with $p=0.3$ & 93.238                                  & (\textcolor{blue}{+0.078})              & 93.805            & (\textcolor{blue}{+0.166}) \\
		SAP with $p=0.4$ & 93.277                                  & (\textcolor{blue}{+0.117})              & 93.811            & (\textcolor{blue}{+0.172}) \\
		SAP with $p=0.5$ & 93.365                                  & (\textcolor{blue}{+0.205})              & 93.861            & (\textcolor{blue}{+0.222}) \\
		SAP with $p=0.6$ & 93.285                                  & (\textcolor{blue}{+0.125})              & 93.752            & (\textcolor{blue}{+0.113}) \\
		SAP with $p=0.7$ & 93.272                                  & (\textcolor{blue}{+0.112})              & 93.730            & (\textcolor{blue}{+0.091}) \\
		SAP with $p=0.8$ & 93.196                                  & (\textcolor{blue}{+0.036})              & 93.743            & (\textcolor{blue}{+0.104}) \\
		SAP with $p=0.9$ & 93.186                                  & (\textcolor{blue}{+0.026})              & 93.546            & (\textcolor{red}{-0.093})  \\
		\bottomrule
	\end{tabular}
	\caption{Test accuracy (\%) for the classification task on the CIFAR-10 dataset, comparing global average pooling (GAP) and stochastic average pooling (SAP). The difference from the baseline performance is presented on the right.}
	\label{tab:c10}
\end{table}

\begin{table}[t!]
	\centering
	\begin{tabular}{l|rr|rr}
		\toprule
		\textbf{Dataset} & \multicolumn{4}{c}{\textbf{CIFAR-100}}                                                                                             \\
		\textbf{Model}   & \multicolumn{2}{c|}{\textbf{ResNet-50}} & \multicolumn{2}{c}{\textbf{ResNet-110}}                                                  \\
		\textbf{Setup}   & \textbf{Accuracy}                       & \textbf{Difference}                     & \textbf{Accuracy} & \textbf{Difference}        \\
		\midrule
		GAP (Baseline)   & 70.582                                  & -                                       & 71.940            & -                          \\
		SAP with $p=0.1$ & 69.865                                  & (\textcolor{red}{-0.717})               & 72.397            & (\textcolor{blue}{+0.457}) \\
		SAP with $p=0.2$ & 70.322                                  & (\textcolor{blue}{-0.260})              & 72.478            & (\textcolor{blue}{+0.538}) \\
		SAP with $p=0.3$ & 70.662                                  & (\textcolor{blue}{+0.080})              & 72.471            & (\textcolor{blue}{+0.531}) \\
		SAP with $p=0.4$ & 70.694                                  & (\textcolor{blue}{+0.112})              & 72.474            & (\textcolor{blue}{+0.534}) \\
		SAP with $p=0.5$ & 70.823                                  & (\textcolor{blue}{+0.241})              & 72.537            & (\textcolor{blue}{+0.597}) \\
		SAP with $p=0.6$ & 70.770                                  & (\textcolor{blue}{+0.188})              & 72.294            & (\textcolor{blue}{+0.354}) \\
		SAP with $p=0.7$ & 70.749                                  & (\textcolor{blue}{+0.167})              & 72.242            & (\textcolor{blue}{+0.302}) \\
		SAP with $p=0.8$ & 70.759                                  & (\textcolor{blue}{+0.177})              & 72.052            & (\textcolor{blue}{+0.112}) \\
		SAP with $p=0.9$ & 70.744                                  & (\textcolor{blue}{+0.162})              & 72.256            & (\textcolor{blue}{+0.316}) \\
		\bottomrule
	\end{tabular}
	\caption{Test accuracy (\%) for the classification task on the CIFAR-100 dataset, comparing global average pooling (GAP) and stochastic average pooling (SAP). The difference from the baseline performance is presented on the right.}
	\label{tab:c100}
\end{table}

We observed that replacing global average pooling with stochastic average pooling improved classification accuracy when using an adequate keep probability. Although certain extreme cases of choosing $p=0.9$ exhibited little difference and $p=0.1$ degraded performance, others with moderate choices of keep probability successfully improved the accuracy. The best performance was found at the middle point $p=0.5$.

Note that PatchDropout opted for lower keep probability such as $p=0.5$ or $p=0.25$ \cite{DBLP:conf/eccv/LinMBHPRDZ14,DBLP:conf/cvpr/HeCXLDG22,DBLP:conf/cvpr/Li0HFH23}. In contrast, for Dropout, a keep probability of $p=0.5$ or $p=0.8$ has been preferred, while the original study on Dropout opted for $p=0.5$ \cite{DBLP:journals/jmlr/SrivastavaHKSS14}. Recent studies \cite{DBLP:conf/icml/TanL19,DBLP:conf/icml/BrockDSS21} reported that sufficient regularization of deep neural networks requires a keep probability of up to $p=0.5$. In theory, Reference \cite{DBLP:conf/nips/BaldiS13} proved that choosing $p=0.5$ yields the highest regularization effect from Dropout in terms of auxiliary weight decay. Based on experimental observations and existing theory, we recommend choosing $p=0.5$ in stochastic average pooling. For the remainder of this paper, we set the keep probability to $p=0.5$.

\subsection{Replace GAP in SE Block}

Besides the classifier head, global average pooling has been used in channel-wise attention modules, such as the SE block. The SE block is a pipeline of [GAP--FC--ReLU--FC--Sigmoid], and we compare the performance before and after replacing its global average pooling with stochastic average pooling. We used SE-ResNet-\{50, 101\} \cite{DBLP:conf/cvpr/HuSS18}, which are variants of ResNet \cite{DBLP:conf/cvpr/HeZRS16} adopting an SE block in each residual block.

We targeted a multi-class classification task on the Oxford-IIIT Pet \cite{DBLP:conf/cvpr/ParkhiVZJ12}, Caltech-101 \cite{DBLP:journals/cviu/Fei-FeiFP07}, and Stanford Cars \cite{DBLP:conf/iccvw/Krause0DF13} datasets. The Oxford-IIIT Pet dataset contains 7K pet images from 37 classes; the Caltech-101 dataset includes 9K object images from 101 classes with a background category; and the Stanford Cars dataset includes 16K car images from 196 classes. These datasets are available on their official websites. Each dataset was split into training, validation, and test sets in a ratio of 70:15:15. All experiments were conducted at a resolution of $224^2$ using standard data augmentation, including random resized cropping to 256 pixels, random rotations within 15 degrees, color jitter with a factor of 0.4, random horizontal flip with a probability of 0.5, center cropping with 224-pixel windows, and mean-std normalization based on ImageNet statistics \cite{DBLP:conf/cvpr/DengDSLL009}.

For training, stochastic gradient descent with a momentum of 0.9, learning rate of 0.1, cosine annealing schedule \cite{DBLP:conf/iclr/LoshchilovH17} with 200 iterations, weight decay of $10^{-3}$, and mini-batch size of 128 were used. These hyperparameters were obtained based on the accuracy of the validation set. The model with the highest validation accuracy was obtained for 200 training epochs, and we report the accuracy on the validation and test sets. The training was conducted on a single GPU machine. An average of three runs with different random seeds was reported for each result.

\begin{table}[t!]
	\centering
	\resizebox{1.0\textwidth}{!}{
		\begin{tabular}{ll|rrr|rrr}
			\toprule
			                                 & \textbf{Backbone} & \multicolumn{3}{c|}{\textbf{SE-ResNet-50}} & \multicolumn{3}{c}{\textbf{SE-ResNet-101}}                                                                                         \\
			\textbf{Dataset}                 & \textbf{Split}    & \textbf{GAP}                               & \textbf{SAP}                               & \textbf{Diff}              & \textbf{GAP} & \textbf{SAP} & \textbf{Diff}              \\
			\midrule
			\multirow{2}{*}{Oxford IIIT-Pet} & Val               & 86.733                                     & 86.913                                     & (\textcolor{blue}{+0.180}) & 87.665       & 88.147       & (\textcolor{blue}{+0.482}) \\
			                                 & Test              & 84.266                                     & 84.717                                     & (\textcolor{blue}{+0.451}) & 85.199       & 86.011       & (\textcolor{blue}{+0.812}) \\
			\midrule
			\multirow{2}{*}{Caltech-101}     & Val               & 81.765                                     & 82.057                                     & (\textcolor{blue}{+0.292}) & 84.221       & 84.488       & (\textcolor{blue}{+0.267}) \\
			                                 & Test              & 80.549                                     & 81.011                                     & (\textcolor{blue}{+0.462}) & 82.713       & 84.051       & (\textcolor{blue}{+1.338}) \\
			\midrule
			\multirow{2}{*}{Stanford Cars}   & Val               & 87.364                                     & 88.326                                     & (\textcolor{blue}{+0.962}) & 86.252       & 87.845       & (\textcolor{blue}{+1.593}) \\
			                                 & Test              & 85.538                                     & 86.499                                     & (\textcolor{blue}{+0.961}) & 84.109       & 85.812       & (\textcolor{blue}{+1.703}) \\
			\bottomrule
		\end{tabular}
	}
	\caption{Accuracy (\%) on validation and test sets using SE-ResNets for the classification task for Oxford IIIT-Pet, Caltech-101, and Stanford Cars datasets. Replacing global average pooling (GAP) within an SE block with stochastic average pooling (SAP) improved performance.}
	\label{tab:se}
\end{table}

\tabref{tab:se} summarizes the results. Compared with the baseline using global average pooling in the SE block, the use of stochastic average pooling improved the performance. These improvements were consistently observed across three datasets and two SE-ResNets, which advocates the use of stochastic average pooling within the SE block for the representation of aggregated information on a feature map.

\subsection{Replace Average Pooling in Semantic Segmentation Networks}

Now, we examine the use of stochastic average pooling on another task. Here we investigate semantic segmentation, which performs pixel-wise classification of images. We targeted PSPNet \cite{DBLP:conf/cvpr/ZhaoSQWJ17} and UPerNet \cite{DBLP:conf/eccv/XiaoLZJS18}, which are representative models in the semantic segmentation task. They include the pyramid pooling module, which uses four 2D average poolings with pooling sizes $r \times r \in \{1^2, 2^2, 3^2, 6^2\}$ to aggregate multi-level feature maps on different contexts. We replaced each average pooling with stochastic average pooling and measured the performance change for semantic segmentation.

The target datasets were the ISPRS Potsdam and Vaihingen \cite{rottensteiner2012isprs}, which contain urban images along with their corresponding segmentation labels. Following the common practice for semantic segmentation of these datasets, a crop size of $512 \times 512$ pixels was used, which was obtained after applying mean-std normalization and a random resize operation using a size of $512 \times 512$ pixels with a ratio range of 0.5 to 2.0. Furthermore, random flipping with a probability of 0.5 and photometric distortions, including brightness, contrast, saturation, and hue, were applied. The objective was to classify each pixel into one of six categories and to train the segmentation network using the cross-entropy loss function.

A training recipe from \texttt{MMSegmentation} \cite{mmseg2020} was employed. For training, stochastic gradient descent with momentum 0.9, weight decay $5 \times 10^{-4}$, and learning rate $10^{-2}$ with polynomial decay with an 80K scheduler were used. Two backbones of ResNet-\{50, 101\} \cite{DBLP:conf/cvpr/HeZRS16} pretrained on ImageNet \cite{DBLP:conf/cvpr/DengDSLL009} were examined. The training was performed on a $4\times$GPU machine, and SyncBN \cite{DBLP:conf/cvpr/0005DSZWTA18} was used for distributed training. We measured three indices commonly used in semantic segmentation---all pixel accuracy (aAcc), mean accuracy of each class (mAcc), and mean intersection over union (mIoU) (Tables~\ref{tab:pot}, \ref{tab:vai}).

\begin{table}[t!]
	\centering
	\begin{tabular}{l|l|rrr|rrr}
		\toprule
		                         & \textbf{Dataset}  & \multicolumn{6}{c}{\textbf{ISPRS Potsdam}}                                                                                                                                \\
		                         & \textbf{Backbone} & \multicolumn{3}{c|}{\textbf{ResNet-50}}    & \multicolumn{3}{c}{\textbf{ResNet-101}}                                                                                      \\
		\textbf{Model}           & \textbf{Setup}    & \textbf{AP}                                & \textbf{SAP}                            & \textbf{Diff}             & \textbf{AP} & \textbf{SAP} & \textbf{Diff}             \\
		\midrule
		\multirow{3}{*}{PSPNet}  & aAcc              & 88.31                                      & 88.32                                   & (\textcolor{blue}{+0.01}) & 88.41       & 88.56        & (\textcolor{blue}{+0.15}) \\
		                         & mIoU              & 73.70                                      & 73.84                                   & (\textcolor{blue}{+0.14}) & 73.78       & 74.30        & (\textcolor{blue}{+0.52}) \\
		                         & mAcc              & 81.95                                      & 82.15                                   & (\textcolor{blue}{+0.20}) & 81.87       & 82.50        & (\textcolor{blue}{+0.63}) \\
		\midrule
		\multirow{3}{*}{UPerNet} & aAcc              & 88.39                                      & 88.51                                   & (\textcolor{blue}{+0.12}) & 88.46       & 88.55        & (\textcolor{blue}{+0.09}) \\
		                         & mIoU              & 73.72                                      & 74.14                                   & (\textcolor{blue}{+0.42}) & 74.04       & 74.28        & (\textcolor{blue}{+0.24}) \\
		                         & mAcc              & 81.94                                      & 82.26                                   & (\textcolor{blue}{+0.32}) & 82.15       & 82.45        & (\textcolor{blue}{+0.30}) \\
		\bottomrule
	\end{tabular}
	\caption{Results (\%) of semantic segmentation on the ISPRS Potsdam dataset before and after replacing average pooling (AP) with stochastic average pooling (SAP). The use of SAP consistently improved the performance.}
	\label{tab:pot}
\end{table}

\begin{table}[t!]
	\centering
	\begin{tabular}{l|l|rrr|rrr}
		\toprule
		                         & \textbf{Dataset}  & \multicolumn{6}{c}{\textbf{ISPRS Vaihingen}}                                                                                                                                \\
		                         & \textbf{Backbone} & \multicolumn{3}{c|}{\textbf{ResNet-50}}      & \multicolumn{3}{c}{\textbf{ResNet-101}}                                                                                      \\
		\textbf{Model}           & \textbf{Setup}    & \textbf{AP}                                  & \textbf{SAP}                            & \textbf{Diff}             & \textbf{AP} & \textbf{SAP} & \textbf{Diff}             \\
		\midrule
		\multirow{3}{*}{PSPNet}  & aAcc              & 89.86                                        & 90.03                                   & (\textcolor{blue}{+0.17}) & 90.04       & 90.09        & (\textcolor{blue}{+0.05}) \\
		                         & mIoU              & 72.27                                        & 72.98                                   & (\textcolor{blue}{+0.71}) & 73.02       & 73.32        & (\textcolor{blue}{+0.30}) \\
		                         & mAcc              & 79.29                                        & 79.94                                   & (\textcolor{blue}{+0.65}) & 80.18       & 80.29        & (\textcolor{blue}{+0.11}) \\
		\midrule
		\multirow{3}{*}{UPerNet} & aAcc              & 89.94                                        & 90.06                                   & (\textcolor{blue}{+0.12}) & 90.05       & 90.14        & (\textcolor{blue}{+0.09}) \\
		                         & mIoU              & 72.66                                        & 72.88                                   & (\textcolor{blue}{+0.22}) & 72.47       & 73.27        & (\textcolor{blue}{+0.80}) \\
		                         & mAcc              & 79.90                                        & 79.93                                   & (\textcolor{blue}{+0.03}) & 79.71       & 80.49        & (\textcolor{blue}{+0.78}) \\
		\bottomrule
	\end{tabular}
	\caption{Results (\%) of semantic segmentation on the ISPRS Vaihingen dataset before and after replacing average pooling (AP) with stochastic average pooling (SAP). The use of SAP consistently improved the performance.}
	\label{tab:vai}
\end{table}

Compared with the baseline that corresponds to using vanilla average poolings, the use of stochastic average pooling consistently improved the segmentation performance. These improvements by stochastic average pooling were observed for all different setups of backbones, segmentation models, datasets, and segmentation indices.

\subsection{Replace GAP in Object Detection Networks}

Now, we target an object detection task, which aims to locate bounding boxes for objects within images. We examined the recent model of DyHead \cite{DBLP:conf/cvpr/DaiCX0LY021}, which uses its own attention mechanism that contains global average pooling. Specifically, a DyHead neck consists of multiple DyHeadBlocks, and each DyHeadBlock contains global average pooling to aggregate information on a feature map. We investigated the performance of DyHead when replacing global average pooling with stochastic average pooling.

For training and testing, we used the COCO 2017 dataset \cite{DBLP:conf/eccv/LinMBHPRDZ14}, which consists of 118K training images, 5K validation images, and 41K test images. Following the common practice for object detection of the COCO 2017 dataset, we applied mean-std normalization, a resize operation using a resize scale of (1333, 800) pixels, and a random flipping with a probability of 0.5. A training recipe from \texttt{MMDetection} was employed. For training, stochastic gradient descent with momentum 0.9, weight decay $10^{-4}$, and learning rate $2 \times 10^{-2}$ with multi-step decay using an $1 \times$ scheduler were used. ResNet-50 \cite{DBLP:conf/cvpr/HeZRS16} pretrained on ImageNet \cite{DBLP:conf/cvpr/DengDSLL009} was employed as the backbone. The training was performed on a $4\times$GPU machine, and SyncBN \cite{DBLP:conf/cvpr/0005DSZWTA18} was used for distributed training. Average precision and its variants, which are commonly used indices, were measured (\tabref{tab:dyhead}).

\begin{table}[t!]
	\centering
	\begin{tabular}{l|cccccc}
		\toprule
		\textbf{Model}  & AP                       & AP$_{50}$                & AP$_{75}$                & AP$_S$                   & AP$_M$                   & AP$_L$                   \\
		\midrule
		DyHead with GAP & 41.7                     & 58.9                     & 45.6                     & 24.8                     & 45.5                     & 54.3                     \\
		DyHead with SAP & 42.1                     & 59.4                     & 45.9                     & 24.9                     & 45.9                     & 54.6                     \\
		Difference      & (\textcolor{blue}{+0.4}) & (\textcolor{blue}{+0.5}) & (\textcolor{blue}{+0.3}) & (\textcolor{blue}{+0.1}) & (\textcolor{blue}{+0.4}) & (\textcolor{blue}{+0.3}) \\
		\bottomrule
	\end{tabular}
	\caption{Results (\%) of object detection on the COCO 2017 dataset before and after replacing global average pooling (GAP) with stochastic average pooling (SAP). AP here exceptionally indicates the average precision index for object detection tasks, not average pooling. Average precision, along with its variants at IoU = 50 and IoU = 75, and for small, medium, and large objects were measured.}
	\label{tab:dyhead}
\end{table}

We observed that the DyHead with stochastic average pooling outperformed that with global average pooling. All these results demonstrate that adopting stochastic average pooling benefits performance compared with vanilla average pooling.

\section{Discussion}
\label{sec:disc}

\begin{figure}[t!]
	\centering
	\includegraphics[width=1.0\textwidth]{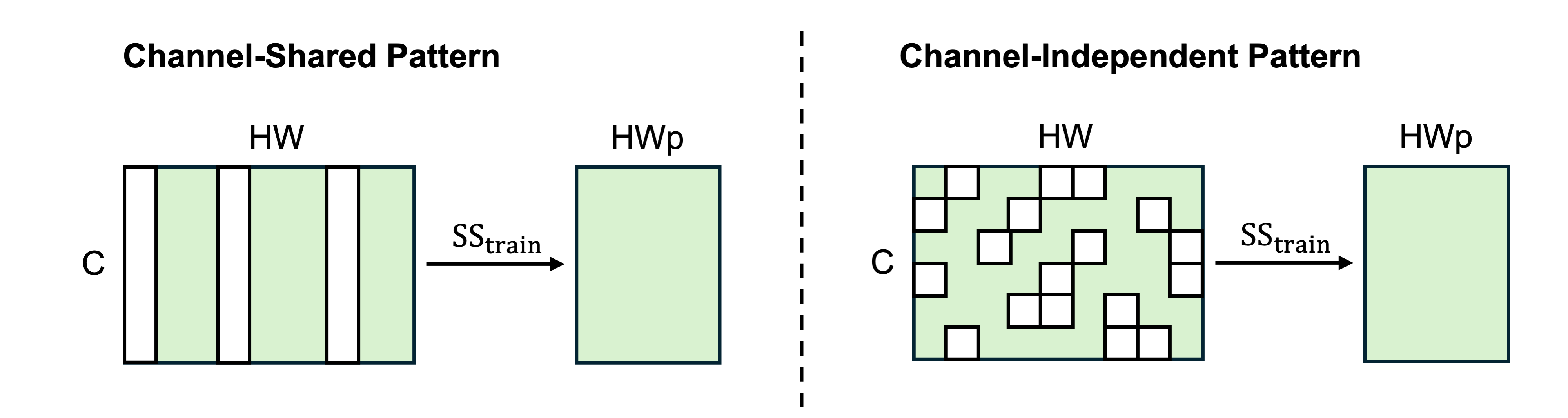}
	\caption{Illustration on subsampling patterns in channel-shared and channel-independent ways}
	\label{fig:channel}
\end{figure}

As mentioned in \secref{sec:sap}, different ways for subsampling may be allowed. In PatchDropout, a channel-shared random pattern is opted for to delete semantic information on the selected patches, whereas a Dropout-like mask applies a fully random pattern that independently subsamples features with respect to the channel dimension (\figref{fig:channel}). Although the different choice of subsampling pattern does not affect our objectives such as matching second moments, it may influence other factors such as preserved information on an image.

\begin{figure}[t!]
	\centering
	\includegraphics[width=1.0\textwidth]{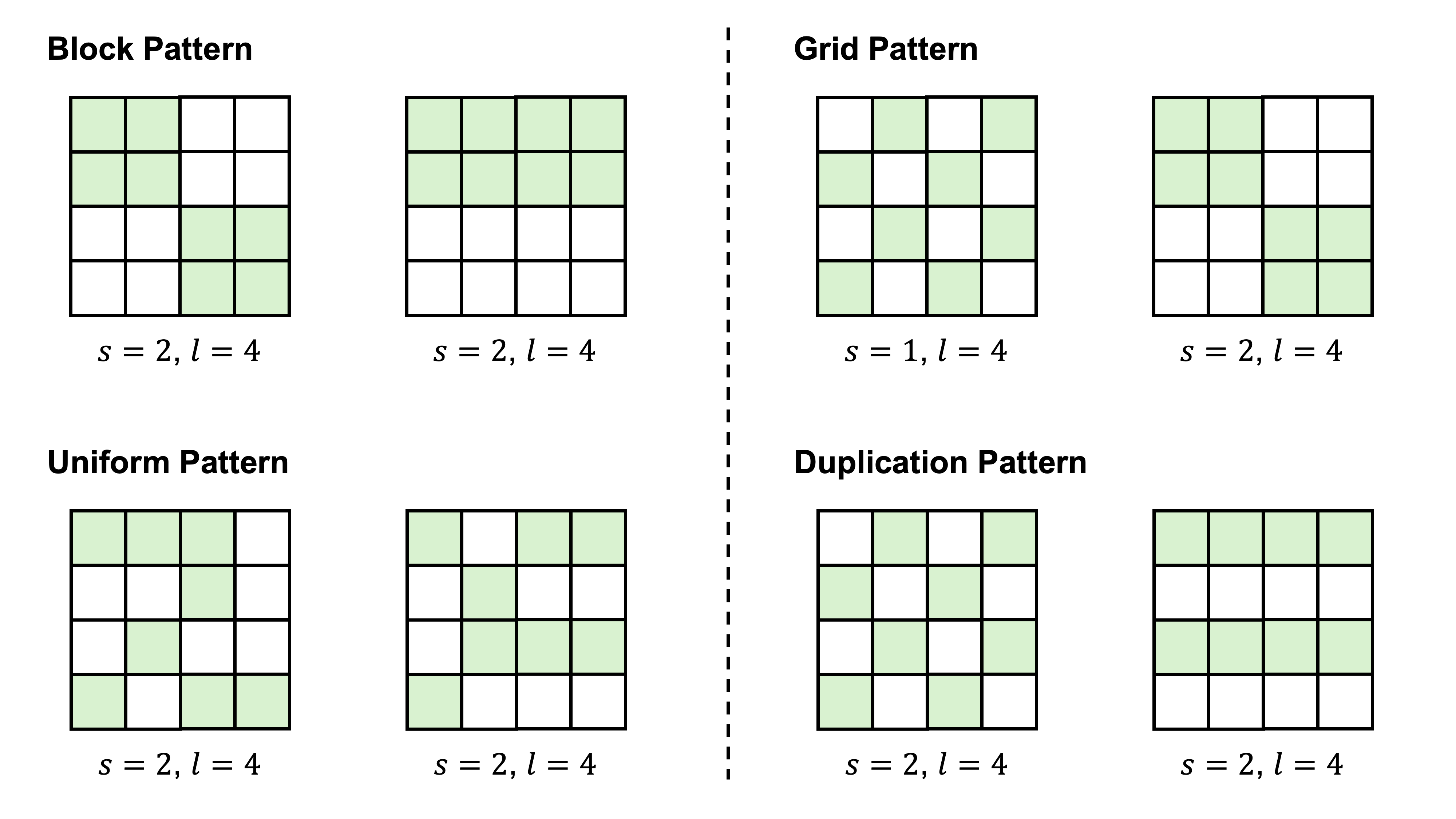}
	\caption{Illustration of the four spatial patterns we examine}
	\label{fig:pattern}
\end{figure}

Furthermore, recent studies such as DropBlock \cite{DBLP:conf/aaai/PhamL21} and AutoDropout \cite{DBLP:conf/nips/GhiasiLL18} reported that adopting specific spatial patterns for Dropout improves performance. In contrast to Dropout, which uses a random pattern without any restriction, the study of DropBlock proposed to drop contiguous features in grouped blocks. They reported that the block pattern readily removes semantic information such as objects or hues in images, which enhances the effect of Dropout in computer vision tasks. The study of AutoDropout employed a reinforcement learning model to search for a novel drop pattern, where they found that a grid pattern worked suitably. The grid pattern facilitates capturing the holistic structure on data, which benefits the effect of Dropout in vision tasks.

Considering these practices for Dropout, here we study adopting a spatial pattern in stochastic subsampling. With the keep probability $p=0.5$ for stochastic subsampling, we consider possible patterns to subsample half of the elements in a feature map. To this end, given an input size $l \times l$ and a factor $s \times s$, four spatial patterns are examined (\figref{fig:pattern}):
\begin{itemize}
	\item Block pattern: The block pattern is inspired by DropBlock \cite{DBLP:conf/aaai/PhamL21}. The whole input area is partitioned into $(l/s) \times (l/s)$ blocks. Each block, with a size of $s \times s$, becomes a group to be subsampled. The pattern without restriction corresponds to $s=1$.
	\item Grid pattern: The grid pattern is inspired by AutoDropout \cite{DBLP:conf/nips/GhiasiLL18}. The subsampling pattern is generated from regular grids, where each grid unit exhibits $s \times s$ size. Because we use the keep probability of $p=0.5$, the stride within a grid is equal to the grid size $s \times s$.
	\item Uniform pattern: The uniform pattern restricts a subsampling pattern to prevent it from being concentrated in a certain area. After partitioning the whole input area into $(l/s) \times (l/s)$ blocks, stochastic subsampling is applied to each block. The pattern without restriction corresponds to $s=l$. The choice of $s=1$ is inapplicable.
	\item Duplication pattern: The duplication pattern is a special case of the uniform pattern and a general case of the grid pattern. After partitioning the whole input area into $(l/s) \times (l/s)$ blocks, a random pattern of size $s \times s$ is sampled as a template. This template is shared across all blocks to form a subsampling pattern. The pattern without restriction corresponds to $s=l$. The choice of $s=1$ is inapplicable.
\end{itemize}
After generating a subsampling pattern from the above list, we additionally apply random circular shifts in vertical and horizontal directions to allow variation in the pattern, preventing fixed partitioning.

Now, we compare the performance of stochastic subsampling when adopting the above patterns. We experimented with the classification task using CIFAR-10 and ResNet-50. The training recipe was the same as the experiment performed in \secref{sec:cifar}. For this task, stochastic average pooling was applied to the last feature map that exhibits a spatial size of $8 \times 8$, \ie, $l \times l=8^2$.

\begin{table}[t!]
	\centering
	\begin{tabular}{l|rr}
		\toprule
		\textbf{Setup}         & \textbf{Accuracy} & \textbf{Difference}        \\
		\midrule
		GAP                    & 93.160            & -                          \\
		SAP (No Restriction)   & 93.365            & (\textcolor{blue}{+0.205}) \\
		\midrule
		Dropout                & 93.109            & (\textcolor{red}{-0.051})  \\
		Channel-Independent    & 93.229            & (\textcolor{blue}{+0.069}) \\
		\midrule
		Block with $s=2$       & 93.236            & (\textcolor{blue}{+0.076}) \\
		Block with $s=4$       & 93.044            & (\textcolor{red}{-0.116})  \\
		\midrule
		Grid with $s=1$        & 93.127            & (\textcolor{red}{-0.033})  \\
		Grid with $s=2$        & 93.190            & (\textcolor{blue}{+0.030}) \\
		Grid with $s=4$        & 93.233            & (\textcolor{blue}{+0.073}) \\
		\midrule
		Uniform with $s=2$     & 93.155            & (\textcolor{red}{-0.005})  \\
		Uniform with $s=4$     & 93.326            & (\textcolor{blue}{+0.166}) \\
		\midrule
		Duplication with $s=2$ & 93.118            & (\textcolor{red}{-0.042})  \\
		Duplication with $s=4$ & 93.191            & (\textcolor{blue}{+0.031}) \\
		\bottomrule
	\end{tabular}
	\caption{Test accuracy (\%) for the classification task on the CIFAR-10 dataset, comparing the effects of adopting different subsampling patterns. SAP (No Restriction) corresponds to the channel-shared pattern.}
	\label{tab:pattern}
\end{table}

\tabref{tab:pattern} summarizes the experimental results. Firstly, we observed that the channel-independent pattern exhibited lower performance compared with the channel-shared pattern. Thus, the channel-shared pattern is more advantageous for removing semantic information from an image to enhance its effect. We also measured performance with standard Dropout, which slightly degraded performance.

Secondly, we measured the performance of using different spatial patterns. In fact, all four patterns failed to exceed the baseline performance of stochastic average pooling that did not apply restriction on the subsampling pattern. Specifically, strong restrictions such as the block pattern with $s=4$ severely degraded accuracy. Rather, weak restrictions such as the uniform pattern with $s=4$ exhibited reasonable performance but still underperformed compared with using a subsampling pattern without restriction. In short, stochastic average pooling worked suitably when allowing sufficient degree of freedom for randomness in the subsampling pattern.

Indeed, because stochastic subsampling follows average pooling in our module, subsampling a unit after average pooling drops all elements within the pooling size. Thus, the effective subsampling pattern with respect to the input already exhibits a grid pattern in stochastic average pooling, which explains why introducing an additional pattern in stochastic subsampling did not improve the performance here. In other words, adopting a spatial pattern causes excessive restrictions on the effective subsampling pattern, which raises another side effect on stochastic average pooling and degrades performance. Based on these observations, we opt for allowing degree of freedom for the randomness of the subsampling pattern in stochastic subsampling, ensuring no restrictions on the spatial pattern.

\section{Conclusion}

This research studied preventing an overfitting problem of deep neural networks by module-level regularization. We pointed out that existing regularization methods such as Dropout and PatchDropout cause inconsistent properties such as second moment and size during training and test phase, which makes them difficult to be used in practical scenarios. To address this issue, we proposed a new module called stochastic average pooling. Stochastic average pooling---which combines stochastic subsampling, average pooling, and $\sqrt{p}$ scaling---ensures consistent properties in output, solving the aforementioned issue. Similarly to Dropout, stochastic average pooling achieves a regularization effect through the ensemble behavior of possible subnetworks. Furthermore, our design for this module enables it to seamlessly replace the existing average pooling architecture. Comprehensive evaluations showed that replacing average pooling with stochastic average pooling consistently improved performance of deep neural networks, which demonstrates its wide applicability.

\bibliography{main}
\bibliographystyle{plain}

\end{document}